\documentclass[journal]{IEEEtran}

\usepackage{silence}
\usepackage[T1]{fontenc}        

\usepackage{moreverb,url}
\usepackage{anyfontsize}
\usepackage[hidelinks]{hyperref}

\usepackage{xcolor}
\usepackage{soul}
\usepackage{graphicx}
\usepackage{dblfloatfix}
\usepackage{tikz}
\usetikzlibrary{arrows.meta,backgrounds,calc,external,patterns,shapes,trees}
\usepackage{tikzscale}
\usepackage{pgfplots}
\pgfplotsset{compat=1.14}
\usepgfplotslibrary{statistics}
\usepackage{subcaption}

\usepackage{booktabs}

\usepackage{amsmath,amssymb}
\usepackage{bm}
\usepackage{siunitx}
\usepackage{algorithm}
\usepackage{algpseudocode}

\usepackage[noadjust]{cite}

\usepackage{textcomp}

\newcommand{\tops}{P}
\newcommand{\preps}{S}
\newcommand{\real}{\mathbb{R}}
\newcommand{\angles}{\mathbb{S}^1}
\newcommand{\posints}{\mathbb{Z}^+}

\newcommand{\springParam}{\Lambda}
\newcommand{\pmParam}{\Theta}

\newcommand{\topfig}{p_f}
\newcommand{\topref}{p_{r_{1 \dots n}}}
\newcommand{\topcon}{p_c}
\newcommand{\prep}{s}

\newcommand{\relclause}[5][math]{
  rel\left(
    \ifstrequal{#1}{text}{\text{#2}}{#2}, \;
    \ifstrequal{#1}{text}{\text{#3}}{#3}, \;
    \{\ifstrequal{#1}{text}{\text{#4}}{#4}\}, \;
    \ifstrequal{#1}{text}{\text{#5}}{#5}
  \right)
}
\newcommand{\relclausetight}[5][math]{
  rel\left(%
    \ifstrequal{#1}{text}{\text{#2}}{#2},
    \ifstrequal{#1}{text}{\text{#3}}{#3},
    \{\ifstrequal{#1}{text}{\text{#4}}{#4}\},
    \ifstrequal{#1}{text}{\text{#5}}{#5}%
  \right)
}

\newcommand{\toponym}{p}
\newcommand{\refframe}[1]{\mathcal{#1}}
\newcommand{\posx}{x}
\newcommand{\posy}{y}
\newcommand{\dist}{r}
\newcommand{\dir}{\theta}

\newcommand{\locclause}[7][math]{
  loc\left(
    \ifstrequal{#1}{text}{\text{#2}}{#2}, \;
    #3, \; #4, \; #5, \; #6, \; #7
  \right)
}
\newcommand{\locclausetight}[7][math]{
  loc\left(%
    \ifstrequal{#1}{text}{\text{#2}}{#2},
    #3, #4, #5, #6, #7%
  \right)
}

\newcommand{\clauses}{c}
\newcommand{\state}{\mathbf{x}}
\newcommand{\dstate}{\dot{\state}}

\newcommand{\parK}{K}
\newcommand{\parXr}{r_n}
\newcommand{\parXt}{\theta_n}

\newcommand{\fnCon}{f}
\newcommand{\fnSpr}{\sigma}
\newcommand{\fnDam}{\tau}
\newcommand{\fnExp}{\lambda}
\newcommand{\fnSet}{\zeta}
\newcommand{\fnCOM}{C}

\newcommand{\setLa}{L_a}
\newcommand{\setLv}{L_v}

\newcommand{\factor}{\alpha}

\newcommand{\explorefactor}{\mathcal{E}}
\newcommand{\explorestep}{\Delta_{\explorefactor}}

\newcommand{\IEEEwarning}{
  \onecolumn
  \thispagestyle{empty}
  \noindent \textcopyright{} 2020 IEEE. Personal use of this material is permitted. Permission from IEEE must be obtained for all other uses, in any current or future media, including reprinting / republishing this material for advertising or promotional purposes, creating new collective works, for resale or redistribution to servers or lists, or reuse of any copyrighted component of this work in other works

  \bigskip \noindent Article published in the \textit{IEEE Transactions on Cognitive and Developmental Systems}: \href{http://doi.org/10.1109/TCDS.2020.2993855}{10.1109/TCDS.2020.2993855}.
  \twocolumn
  \newpage
  \pagenumbering{arabic}
}

\newcommand{\shrinkfactor}{-0.5em}

\begin{document}

\IEEEwarning

\title{Robot Navigation in Unseen Spaces using an Abstract Map}
\author{Ben Talbot, Feras Dayoub, Peter Corke, \textit{Fellow, IEEE}, and Gordon Wyeth, \textit{Member, IEEE}%
  \thanks{The authors are with the School of Electrical Engineering and Computer Science, Queensland University of Technology (QUT), Brisbane, Australia. (email: b.talbot@qut.edu.au)}%
}

\maketitle

\begin{abstract}
  Human navigation in built environments depends on symbolic spatial information which has unrealised potential to enhance robot navigation capabilities. Information sources such as labels, signs, maps, planners, spoken directions, and navigational gestures communicate a wealth of spatial information to the navigators of built environments; a wealth of information that robots typically ignore. We present a robot navigation system that uses the same symbolic spatial information employed by humans to purposefully navigate in unseen built environments with a level of performance comparable to humans. The navigation system uses a novel data structure called the \textit{abstract map} to imagine malleable spatial models for unseen spaces from spatial symbols. Sensorimotor perceptions from a robot are then employed to provide purposeful navigation to symbolic goal locations in the unseen environment. We show how a dynamic system can be used to create malleable spatial models for the abstract map, and provide an open source implementation to encourage future work in the area of symbolic navigation. Symbolic navigation performance of humans and a robot is evaluated in a real-world built environment. The paper concludes with a qualitative analysis of human navigation strategies, providing further insights into how the symbolic navigation capabilities of robots in unseen built environments can be improved in the future.
\end{abstract}

\begin{IEEEkeywords}
  symbol grounding, symbolic spatial information, abstract map, navigation, cognitive robotics, intelligent robots
\end{IEEEkeywords}

\section{Introduction}

\begin{figure}[p]
  \centering
  \input{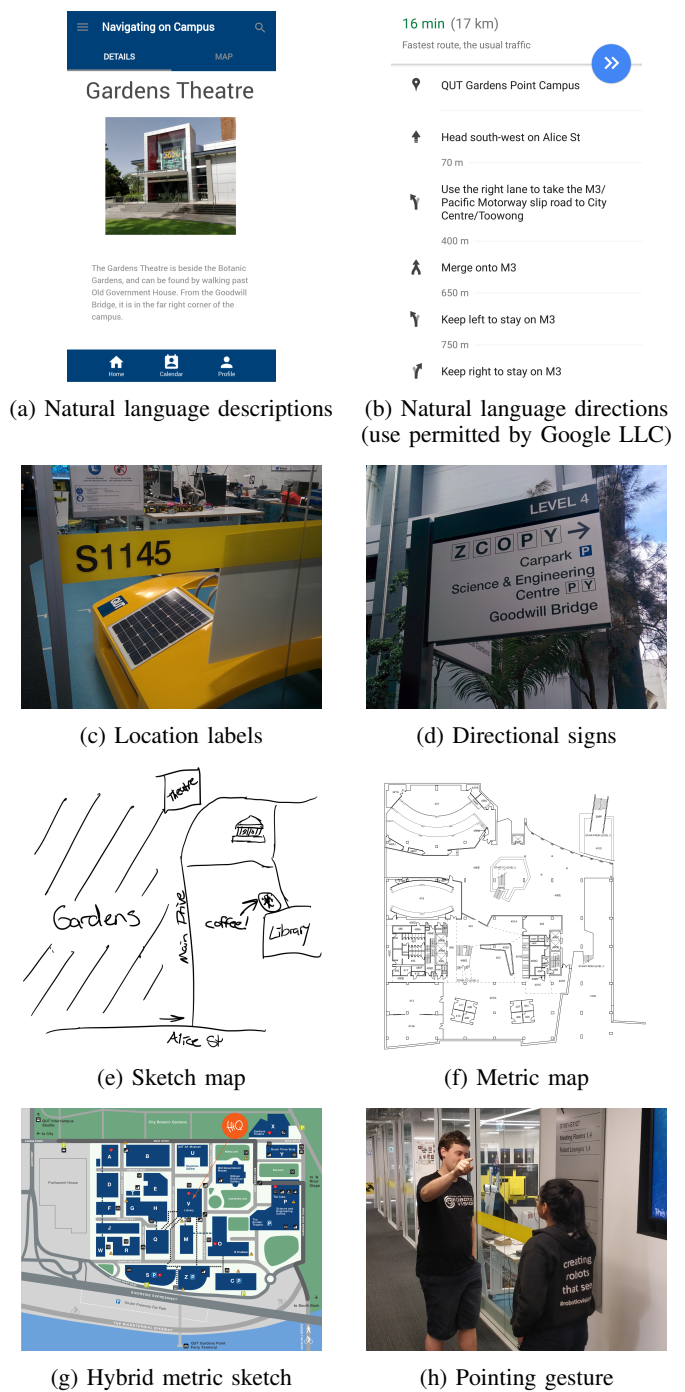}
  \caption{Examples of different types of navigation cues and symbolic methods used to describe spatial relations.}
  \label{fig:navigation_cues}
\end{figure}

Proficiently navigating through unseen urban environments is a vital part of daily life for humans, whether it be meeting in a new colleague's office, making it to the correct gate for a plane in a foreign airport, locating an apartment while on overseas holiday, finding the lion at the zoo, or even finding bananas in a new grocery store. Robots must develop the same navigation abilities which humans exhibit if they are to truly become useful co-inhabitants of built environments.

Wayfinding \cite{mollerup2013wayshowing}, the human navigation process in built environments like offices or shopping centres, relies on a type of spatial cue called a \textit{navigation cue}. Navigation cues come in many forms including labels, signs, maps, planners, structural landmarks, spoken directions, and navigational gestures. A subset is shown in Fig. \ref{fig:navigation_cues}. Navigation cues embed rich spatial information, and are placed throughout an environment to aid the navigation of people who have never been there before (e.g. labels are placed on the outside of offices, floor plans and maps at main entrances, and signs at choice points in corridors or walkways). Wayshowing \cite{permollerup2005} is a set of purposeful design practices and principles that inform the placement of navigation cues so as to maximise environment navigability. Wayshowing plays a key role in the architectural design of built environments.

Navigation cues provide a special class of navigation information referred to as \textit{symbolic spatial information} to convey information about the spatial structure of the world. Symbols are the backbone of human communication with simple elements such as words, phrases, pictures, arrows, and gestures employed to concisely represent spatial concepts. The conciseness in symbolic representations is achieved by omitting superfluous details, instead relying on the observer's capabilities and experiences to decode symbolically communicated concepts while deducing missing details. The nature of symbols often results in symbolic spatial information being ambiguous, or challenging to correlate with meaning in the real world \cite{landau1993whence}. Nevertheless humans can capably and effortlessly leverage the richness of symbolic spatial information to profound effect.

In contrast, robots are typically oblivious to the rich spatial information available in navigation cues. Robots instead use raw, low-level sensorimotor measurements to navigate their environments. The measurements typically come in the form of either range and bearing data for surrounding obstacles \cite{durrant2006simultaneous,montemerlo2002fastslam,grisetti2007improved}, or snapshots of visual appearance \cite{milford2004ratslam}. Navigation is then performed using spatial models and algorithms representing the geometric structure of a space, with no incorporation of semantics. Robots that navigate using only low-level sensor measurements are incapable of purposeful navigation when applied in spaces previously unvisited by the robot. Such spaces are referred to in this paper as \textit{unseen spaces}.

Symbolic spatial information provides an opportunity to create richer spatial models than those solely estimating geometric structure from a robot's sensor measurements. Robotic systems that use symbols to navigate are not prevalent in the literature, and those that exist carry varying restrictions. Such restrictions include requiring human-constructed spatial models \cite{fasola2013using}; inferring semantics solely from object occurrences in spaces \cite{galindo2008robot,kollar2010toward}; probabilistic models limited to seen spaces \cite{walter2013learning,tellex2011approaching}; and using limited symbol sets \cite{bauer2009autonomous,elmogy2011multimodal,boniardi2016autonomous} like pointing in unseen spaces. The utility of symbols for robot navigation in built environments remains untapped.

\begin{figure}[t]
  \centering
  \input{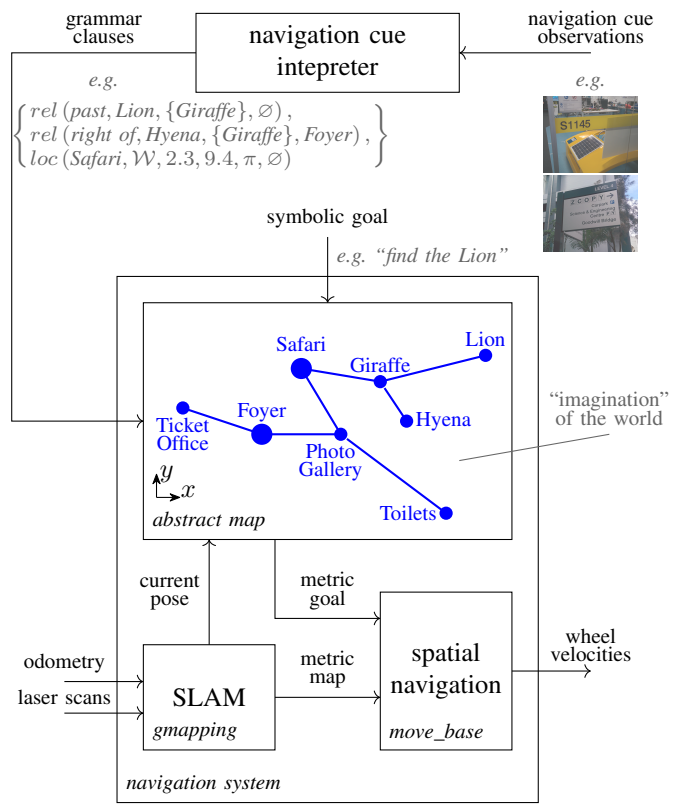}
  \caption{System diagram for a navigation system using navigation cue observations to navigate an unseen space. The abstract map uses a malleable spatial model to tether spatial symbols to direct robot perceptions.}
  \vspace{\shrinkfactor}
  \label{fig:system_outline}
\end{figure}

We present a navigation system that leverages both the abstract nature of navigation symbols and traditional geometric spatial models to provide purposeful navigation in unseen built environments. The system employs a malleable spatial model called the \textit{abstract map} shown in Fig. \ref{fig:system_outline}. It allows a traditional robot navigation system to utilise the symbolic spatial information embedded in navigation cues. Our research provides the following contributions in the area of symbolic navigation for mobile robots:
\begin{itemize}
  \item a robotics-oriented grammar, with hand-crafted clauses, used to express the spatial information communicated by navigation cues (the perceptual challenges associated with extracting symbolic spatial information from images of navigation cues are left as open research questions);
  \item the abstract map, a malleable spatial model used by the robot navigation system to purposefully navigate unseen spaces;
  \item a novel method for using a dynamic multi-body system to ``imagine'' malleable spatial models of unseen built environments;
  \item procedures for reconciling spatial models imagined from symbols with information received from the direct sensorimotor perceptions of the robot; and
  \item an open source implementation of the abstract map---available at \url{https://btalb.github.io/abstract_map/}
\end{itemize}
The abstract map is evaluated in a study comparing robot to human navigation performance in an unseen real built environment. We present the following findings from the study:
\begin{itemize}
  \item a quantitative comparison of human and abstract map navigation performance,
  \item qualitative insights into the human navigation process, and
  \item suggestions as to how robot symbolic navigation systems can be improved in the future.
\end{itemize}

The rest of the paper is organised as follows. Section \ref{sec:related} describes the use of symbols in navigation cues and robot navigation systems. The abstract map is formally defined in Section \ref{sec:abstract_map}, with the experimental procedure and results then presented in Sections \ref{sec:procedure} and \ref{sec:results} respectively. The paper concludes in Section \ref{sec:conclusion} with a discussion of the results, and suggestions for future work.

\section{Background and Related Work}
\label{sec:related}

To understand how a robot can use symbolic spatial information to inform its navigation process, there are three relevant topics in the literature: 1) how navigation cues communicate symbolic spatial information, 2) robotic interpretations of the symbol grounding process, and 3) the use of symbols in the robot navigation process. Each of these are explored in detail in the sections below.

\subsection{Symbolic Spatial Information from Navigation Cues}
\label{subsec:cues}

Symbols are central in every navigation cue that humans place in their built environments. The diversity of symbols employed in navigation cues is large: arrows are used for signboards; arbitrary labels exist for roads, train stations, buildings, offices, etc.; words are used to communicate spatial directions; pictorial artefacts are used in maps and sketches; and even a basic action like pointing a finger can be used to signify direction. Navigation cues use symbols for two distinct purposes: to name a location in the world, and to describe a spatial relationship between locations.

When referring to locations in the real world, a linguistic symbol called a \textit{toponym} is used. Toponyms---also referred to as labels, locations, places, or spaces \cite{schulz2009spatial}---are nouns used to refer to any classification of space, typically encapsulated by some form of basic geometric structure. The geometric structure can be a point (e.g. corner of Main Street and First Street), a one-dimensional path (e.g. Main Street), a region of a two-dimensional plane (e.g. block 37 on Main Street), or a three-dimensional volume (e.g. Sciences Building) \cite{tversky2003places}.

The second use of symbols in navigation cues---describing spatial relationships between locations---employs a much wider range of symbol types, with intrinsic elements of the navigation cue often playing a crucial role in the symbolic communication. For example, the arrow symbol on a directional sign requires the observer to use the location and orientation of the sign in the real world to interpret the symbol. Symbolic methods for describing spatial relationships are split into four distinct types, which are discussed in detail in the following paragraphs (examples of these are shown in Fig. \ref{fig:navigation_cues}).

\subsubsection{Natural language descriptions and directions}

are examples of linguistic navigation cues, which can be either spoken or written. Natural language descriptions use linguistic symbols to describe the spatial relationship between toponyms. Sequential directions additionally use ordering to break a complex path into a sequence of spatial relations. Examples of linguistic cues can be seen in Fig. \ref{subfig:cues_descriptions} and \ref{subfig:cues_directions}.

Linguistic navigation cues use a set of words called \textit{spatial prepositions} \cite{tyler2003semantics}---a subset of only 80 to 100 prepositions in the entire English language \cite{landau1993whence}---to describe the spatial relationship between spaces. Examples include ``left'', ``right'', ``towards'', ``beside'', ``between'', ``past'', etc. Interpretation of spatial prepositions uses simple units of space, with no more geometric complexity required than points, containers, volumes, or units with basic axial structure (like a tree with nodes and edges) \cite{landau1993whence,van2003representing}.

Spatial prepositions describe the spatial relationship of a target called the \textit{figure}, relative to a reference location called the \textit{reference object}. Fig. \ref{fig:linguistic_components} shows how prepositions and toponyms typically combine in phrases, with an included or implied \textit{context} playing an influential role in the interpretation of a spatial relationship \cite{levinson2003space}. Sequential directions for instance, assume the context is where the last step finished.

\begin{figure}[t]
  \centering
  \includegraphics{./figs/linguistic_components.tikz}
  \vspace{-1ex}
  \caption{The key components of a linguistic cue: a spatial preposition describes the location of a figure with respect to a reference object, with context often aiding in interpretation (e.g. interpreting ``left of'').}
  \vspace{\shrinkfactor}
  \label{fig:linguistic_components}
\end{figure}

\subsubsection{Labels and signboards}

are examples of locational cues, which communicate a spatial relation to an observer through their location in the environment. Label cues mark the real-world location of a toponym, whereas directional signs use arrows and approximate distances to describe a toponym's location relative to the cue's real-world location. Example locational cues can be seen in Fig. \ref{subfig:cues_labels} and \ref{subfig:cues_signs}.

Locational cues, long identified as a crucial influence on human wayfinding performance \cite{weisman1981evaluating,o1991effects}, associate places in the real world to symbols embedded in the environment. The association provided is crucial in allowing a navigator to link their internal spatial concepts about the world with what they observe in the environment.

\subsubsection{Sketch maps and metric maps}

are examples of pictorial cues, which use the visual space in a picture to represent spatial concepts \cite{tversky2003structures}. Pictorial cues are classified by how visual space is used to communicate spatial relations. Sketch maps forgo unimportant information to focus solely on emphasising spatial relationships between places. Alternatively, metric maps express spatial relationships using geometric quantities in a to-scale picture. Examples of each type of cue, and a hybrid of both, can be seen in Fig. \ref{subfig:cues_sketch}, \ref{subfig:cues_metric}, and \ref{subfig:cues_metric_sketch} respectively.

Humans find sketch maps a significantly more effective navigation cue than scaled metric maps \cite{wang2012empirical} due to the likeness of their spatial descriptions to human mental structures \cite{tversky2003structures} and approaches \cite{denis1997description}. Consequently, sketch maps can be considered similar to linguistic cues that use pictures in place of toponyms and spatial prepositions. Conversely, metric maps can be considered similar to locational cues but with the extra mental burden of conversion from the map's coordinate frame to the real world.

\subsubsection{Navigational gestures}

communicate spatial information through gestures, a symbolic communication performed through hand movements. Gestures come in four different types \cite{allen2003gestures}---iconics, metaphorics, deictics, and beats---with only iconics and deictics employed in navigation cues. An example iconic gesture is placing a hand in front of the other to visually support the description ``the coffee shop is in front of the building''---the hands are being used as icons for the places. Deictics are the pointing gestures used by speakers to orient the listener in referential space. A common example is the pointing gesture to communicate ``the coffee shop is over there'', as shown in Fig. \ref{subfig:cues_gesture}.

Both forms of navigation gesture can be considered hand-based versions of previously discussed navigation cues. A deictic gesture is a locational cue with greater flexibility in direction than printed arrows (using a sign hanging vertically on a wall to communicate \ang{61} east of north is infeasible), and the cue provider can also move throughout the environment. Iconic gestures are indecipherable without the accompanying linguistic description, and consequently can be thought of as linguistic cues with added verbosity via hand movements.

\subsection{The Symbol Grounding Problem}

\begin{figure}[t]
  \centering
  \includegraphics[width=0.8\columnwidth]{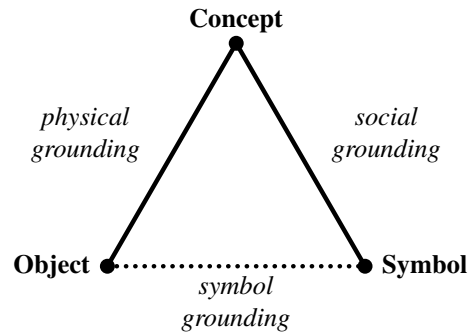}
  \vspace{-1ex}
  \caption{The semiotic triangle {\cite{peirce1902logic}} describes symbol grounding as an outcome of both deriving internal concepts for physical objects, and linking shared symbols to these internal concepts.}
  \vspace{\shrinkfactor}
  \label{fig:semiotic_triangle}
\end{figure}

The core challenge in using symbolic spatial information, for both humans and robots, revolves around extracting meaning from symbols; a problem referred to as the \textit{symbol grounding problem} \cite{harnad1990symbol}. Both robots and humans represent the world through their own internal concepts. However they require a method of representing symbols in terms of their own internal concepts before they can extract real world meaning from symbols. The problem is often represented through the semiotic triangle \cite{ogden1923meaning} (see Fig. \ref{fig:semiotic_triangle}), coined by Peirce \cite{peirce1902logic}.

The semiotic triangle frames symbol grounding as a combination of physical grounding and social grounding. Physical grounding is the linking of internal concepts to the real world \cite{roy2005semiotic,brooks1990elephants,vogt2002physical} whereas social grounding is linking shared concepts, like symbols, to internal concepts \cite{steels2000aibo,schulz2011lingodroids}. In the scope of mobile robotics, transforming sensor data into internal spatial models like maps and pose graphs is considered physical grounding. Once a physical grounding is established, the robot can use these internal spatial models to complete navigation tasks in the real world.

Social grounding in robotics is a process which gives the robot the ability to understand and communicate in a symbolic lexicon. One example is in the emergence of symbols amongst communicating robots. Studies use activities called language games \cite{roy2005semiotic,steels2008symbol} to develop and communicate a shared semiotic symbolic lexicon amongst robot populations \cite{steels2015talking,cangelosi2001evolution,vogt2007social,schulz2011lingodroids}. All of these studies focused on attaching, communicating, and interpreting symbols already linked to robot concepts rather than imposed symbols like human language.

\subsection{Use of Human Symbols in Robot Navigation}

The scale of the human symbolic lexicon, and lack of a universal solution to the symbol grounding problem, makes using humans symbols in robotics a challenging task. As a result, robotic systems typically employ a restrictive subset of human symbols (e.g. only pointing gestures, data structures requiring manual annotation by humans, or limiting language to a handful of words with static interpretations). Additionally, robot navigation typically limits the application of symbols to already explored spaces. However, utilising symbols only in observed spaces misses the fundamental utility of symbols---sharing human spatial perceptions with robots to enable navigating without requiring prior perception.

Approaches in the literature have advanced from requiring a human-in-the-loop, to automatically linking symbols and robot spatial models. Human-in-the-loop approaches have progressed from using humans to interpret and follow automatically generated navigation instructions \cite{sriharee2013indoor}, to robots using human-annotated semantic maps to follow complex natural language instructions like ``go to the kitchen while hugging the right wall'' \cite{fasola2013using}. Significant progress has been demonstrated in the area of semantic mapping {\cite{kostavelis2015semantic}}, with approaches for attaching symbols to maps including using discrete areas of segmented robot maps \mbox{\cite{galindo2005multi,galindo2008robot}}, linking perceptions with ontological representations \mbox{\cite{kostavelis2016robot,pronobis2012large}}, and probabilistically inferring symbolic labels from object detections (e.g. attaching the ``office'' symbol to a space because computers and desks were detected) {\cite{kollar2010toward}}. Work using probabilistic inference has culminated in the generalised grounding graph (GGG) \cite{tellex2011approaching} and extensions \cite{howard2014natural,chung2015performance}, which interprets novel commands by mapping between words in language and concrete objects, places, paths, and events in the external world. Although the progress is significant, demonstrated systems still only consider the scope of spaces already observed by a robot.

Progress in unseen spaces has relied on using constrained subsets of human symbols, or on limiting how far the robot can explore outside of already seen spaces. The guarantee of sequence in route instructions has been exploited with symbol subsets ranging from pointing gestures \cite{bauer2009autonomous} and input restricted to artificial instruction sets like ``\$GO()'' and ``\$FOLLOW()'' \cite{elmogy2011multimodal}, all the way to natural language \cite{macmahon2006walk} and free-hand sketches \cite{boniardi2016autonomous}. A limited semantic vocabulary consisting of four prepositions has been used to improve existing navigation performance in observed spaces \cite{walter2013learning,hemachandra2014learning}. Extensions of the approach use a single novel instruction to find an unseen place, like finding the kitchen using ``go to the kitchen that is down the hallway'' \cite{hemachandra2015learning,duvallet2016inferring}. However, the literature offers no further progress in using symbols in navigating unseen spaces.

Previous work by the authors used symbols from particular types of navigation cues to navigate unseen spaces. Firstly, an abstract map for converting locational cues between the frames of reference of a floor plan and a robot was presented \cite{schulz2015robot}. Next, we demonstrated an abstract map using structured linguistic navigation cues with limited graph-based support for navigating between different spaces \cite{talbot2016find,talbot2015reasoning,talbot2018integrating}. In this work, we expand the scope to address how an abstract map can generically employ the symbolic spatial information embedded in all types of navigation cues for robot navigation in built environments.

\section{The Abstract Map}
\label{sec:abstract_map}

Three concurrent processes are used with the abstract map to harmonise and exploit symbolic and metric spatial information. Firstly, the clauses of a robotics-oriented grammar are used to generically represent the symbolic spatial information embedded in navigation cues. Next, spatial models for unseen spaces are imagined from symbols alone using malleable interpretations of the symbolic spatial information in clauses. Finally, the malleable elements of the abstract map are adapted to reflect the real world perceptions of the robot. In this work, simulated spring dynamics are used to construct the abstract map's malleable spatial model. Each of the processes is described in detail below.

\subsection{Capturing Symbolic Spatial Information from Navigation Cues}

We define a robotics-oriented grammar where sets of clauses represent collections of symbolic spatial information. Clauses consist of real numbers $\real$, angles $\angles$, reference frames, elements of the set of toponyms $\tops$, and elements of the set of spatial prepositions $\preps$. The grammar concisely describes the symbolic spatial information embedded in navigation cues. In this work we do not address the perceptual challenges associated with \textit{how} symbolic spatial information can be extracted from navigation cues, although we have conducted preliminary studies in this area \cite{schulz2015robot,lam2014text,lam2015automated}.

A \textit{relational clause} describes the spatial relationship between symbolic locations and is parameterised by the function
\begin{equation}
  \relclause{\prep}{\topfig}{\topref}{\topcon}
  \label{eq:relational_clause}
\end{equation}
where $\prep \in \preps$ is the preposition used to describe the spatial relationship between the figure toponym $\topfig \in \tops$ and one or more referent toponyms $\topref \in \tops$, given the context toponym $\topcon \in \{\varnothing, \tops\}$ ($\varnothing$ denotes no provided context). The conversion of a navigation cue can produce numerous clauses, with this being formally defined as the conversion to a set of clauses. For example, a set containing a single clause captures the symbolic spatial information in the linguistic cue ``Isla's office is between the entryway and printer'':
\begin{equation*}
  \big\{\relclause[text]{between}{Isla's office}{entryway, printer}{$\varnothing$}\big\}\,.
  \label{eq:relational_clause_example}
\end{equation*}

A \textit{locational clause} links symbolic locations to locations in an environment and is parameterised by the function
\begin{equation}
  \locclause{\toponym}{\refframe{F}}{\posx}{\posy}{\dist}{\dir}
  \label{eq:locational_clause}
\end{equation}
where $\toponym \in \tops$ is the toponym whose location is described as a distance of $\dist \in \{\varnothing,\real\}$ and direction of $\dir \in \{\varnothing,\angles\}$ from the point $(\posx,\posy)$ in reference frame $\refframe{F}$. Here $\varnothing$ denotes that the value can be unspecified. For example an office label for ``Riko's Office'' observed at coordinates $(5.21,1.76)$ relative to the robot would be captured by the set of clauses:
\begin{equation*}
  \big\{\locclause[text]{Riko's Office}{\refframe{W}}{5.21}{1.76}{0}{\varnothing}\big\}
  \label{eq:locational_clause_example}
\end{equation*}
where $\refframe{W}$ is the world frame of reference, $\dist$ is $0$ as the label specifies where a place is, and $\dir$ is unspecified.

Fig. \ref{fig:grammar_examples} shows examples of how the two types of clause can be employed to capture symbolic spatial information from human navigation cues.

\begin{figure}[p]
  \centering
  \input{./figs/cue_clauses.tex}
  \caption{Examples of using a set of clauses from the robotics-oriented grammar to capture the symbolic spatial information communicated by navigation cues.}
  \label{fig:grammar_examples}
\end{figure}

\subsection{Generating Spatial Models from Relational Clauses}
\label{subsec:relational}

\begin{figure}[t]
  \centering
  \includegraphics[width=\columnwidth]{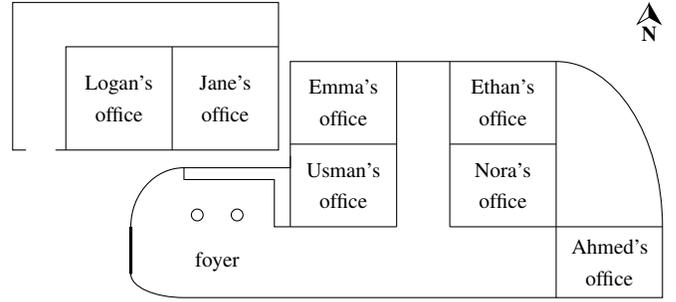}
  \caption{A hypothetical university environment: ``A Block'' is in the top left, and ``B Block'' is on the bottom right.}
  \vspace{\shrinkfactor}
  \label{fig:example_environment}
\end{figure}

Prepositions encoded in relational clauses symbolically describe two spatial properties: spatial layout and spatial hierarchy. Prepositions that describe spatial layout include ``left'', ``down'', ``west'', and ``beside'', whereas the prepositions ``in'', ``contains'', and ``within'' are examples describing spatial hierarchy. Descriptions of layout and hierarchy can be used to inform the imagination of plausible spatial models for unseen spaces. The example built environment shown in Fig. \ref{fig:example_environment} is used below to describe the process undertaken in creating the abstract map's malleable spatial model from relational clauses alone.

\begin{figure}[t]
  \centering
  \input{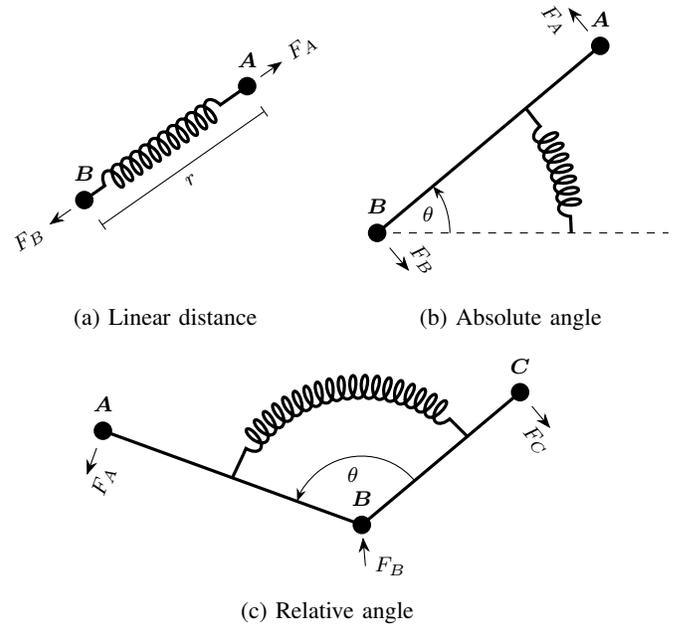}
  \caption{The springs used in imagined spatial models to represent the geometric constraints suggested by relational clauses.}
  \vspace{\shrinkfactor}
  \label{fig:springs}
\end{figure}

Spatial models are created from the symbols in relational clauses by translating clauses into spatial artefacts that capture both the spatial suggestion and malleability inherent in symbol interpretations. Relational clauses are represented in a dynamics-based spatial model by defining toponyms as point-masses which move within a plane, and mapping spatial prepositions to instances of the simulated springs in Fig. \ref{fig:springs}.

Point-mass $i$ in a system is represented by a set of parameters $\pmParam_i$ and state vector $\bm{\xi}_i$. The parameter set $\pmParam_i$ contains toponym $p \in \tops$ and a constant unit mass. The state vector, with respect to the world frame $\refframe{W}$, is defined as
\begin{equation}
  \bm{\xi}_i = \left[\begin{array}{c}\bm{x}_i\\\bm{\dot{x}}_i\end{array}\right] \, : \, \bm{x}_i = \left[\begin{array}{c}x_i\\y_i\end{array}\right] \, : \, x_i,y_i \in \real \,,
  \label{eq:point_mass_state}
\end{equation}
with $\iota(\toponym) \rightarrow i : i \in \posints$ a function mapping toponym $\toponym$ to its index in the set of point-masses $\pmParam$.

\begin{figure*}
  \centering
  \includegraphics[width=\textwidth]{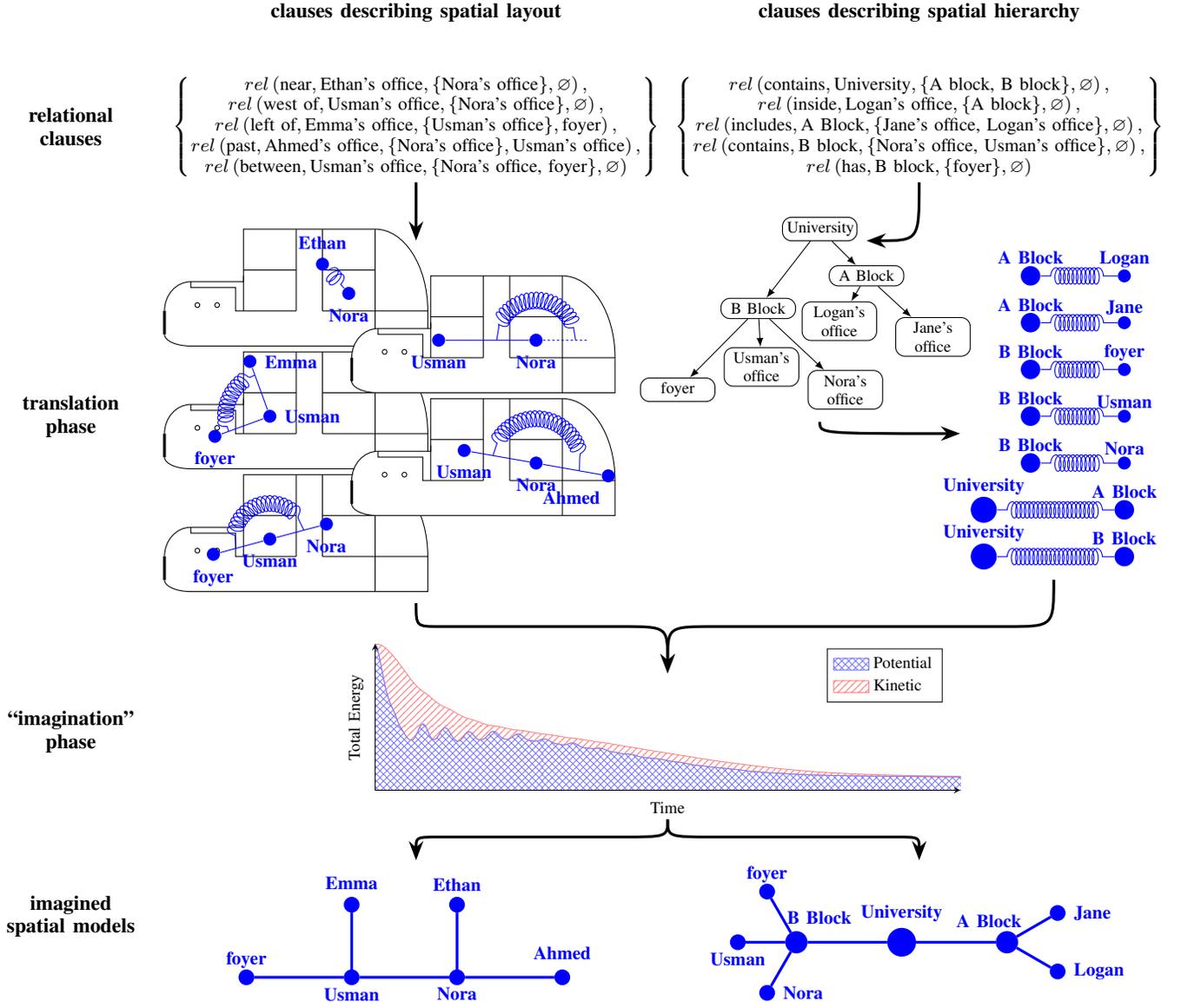}
  \caption{Malleable spatial models for both spatial layout and spatial hierarchy (bottom row) are imagined for the hypothetical university environment from the respective relational clauses (top row). Models for spatial layout and hierarchy are split for illustrative purposes, but exist as a single model in the real system.}
  \label{fig:example_process}
\end{figure*}

Spatial prepositions are mapped to one or more simulated springs which constrain either distance, absolute direction, or relative direction between two or more toponyms. The function $\fnSpr(\mathbf{x},\springParam_j)$ defines the force applied to the system's point-masses by spring $j$, which is defined by the spring parameter $\springParam_j$. $\springParam_j$ consists of the spring type, the toponyms the spring connects to, stiffness $\parK$, and either natural length $\parXr \in \real$ or angle $\parXt \in \angles$. For instance, the preposition ``right of'' is represented by a relative angle spring shown in Fig. \ref{subfig:spring_rel} with natural angle $\parXt = 90\si{\degree}$ between point-masses for the figure toponym and context toponym, relative to the referent toponym. The figure, referent, and context toponyms correspond to nodes A, B, and C  respectively in Fig. \ref{subfig:spring_rel} for this example. The spring is given a moderate stiffness $\parK$ to represent that ``right of'' can apply to scopes outside of precisely orthogonal. Fig. \ref{fig:example_process} demonstrates some more example conversions from symbols to springs in the translation phase.

Spatial hierarchy---when a space is inside another like ``the foyer is in B Block''---is also modelled with springs. Hierarchy suggestions are first added to an evolving directed graph of spatial hierarchy as shown at the start of the translation step in Fig. \ref{fig:example_process}. For example, ``the University contains A Block'' would add a parent-child edge to the graph from the ``University'' node to ``A Block''. Each edge $k$ of the hierarchical graph is then converted to a distance spring $\springParam_k$ with a natural length $\parXr \in \real$ corresponding to the typical distance between spaces at that level of the graph. A very low spring stiffness $\parK$ is used to reflect the sweeping assumptions made in estimating distance solely from spatial hierarchy, and the wide variance of values in reality.

\subsection{Imagining Spatial Models for Unseen Places using Spring-Mass Dynamics}

Simulated dynamics are used to create malleable spatial models for unseen places from the spatial suggestions provided by relational clauses. A spatial model is defined as the position states of each of the $m$ point masses in a system whose state vector is
\begin{equation}
  \mathbf{x} = \left[ \> \bm{\xi}_1, \> \bm{\xi}_2, \> \dots \> , \bm{\xi}_m \right]^\intercal \enspace .
  \label{eq:system_state}
\end{equation}
As shown in Algorithm \ref{alg:imagining}, a spatial model is imagined by performing numerical integration on a starting system state $\state_0$ using the system motion model $\dstate = \fnCon(t,\state)$, until a settling criteria $\fnSet(\dstate)$ is met. The following paragraphs define the process for augmenting a spatial model with new clauses, the components of the motion model $\fnCon(t,\state)$, and the settling criteria $\fnSet(\dstate)$.

\begin{algorithm}[t]
  \textbf{Input:} $\> \state_0, \> \fnCon()$
  \begin{algorithmic}[1]
    \State $\state \gets \state_0$
    \While{\textbf{not} $\fnSet(\dstate)$}
    \State $\dstate \gets \fnCon(t, \state)$
    \State $t, \state \gets \text{odeIntegrate}(t, \Delta{}t, \state, \dstate)$
    \EndWhile
  \end{algorithmic}
  \caption{``Imagining'' a spatial model using point-masses}
  \label{alg:imagining}
\end{algorithm}

\begin{algorithm}[t]
  \textbf{Input:} $\> \clauses_\text{new}, \> \springParam, \> \pmParam$
  \begin{algorithmic}[1]
    \State $\springParam \gets \text{clausesToSprings}(\clauses_\text{new}) \cup \springParam$
    \State $\pmParam_\text{new} \gets \text{pointMassesInSprings}(\springParam) - \pmParam$
    \State $\pmParam_\text{new} \gets \text{sortByTotalStiffness}(\pmParam_\text{new}, \springParam)$
    \For{$p \gets \text{pointMassesToToponyms}(\pmParam_\text{new})$}
    \State $i \gets \iota(p)$
    \State $\bm{\xi}_i \gets \left[ \tilde{x}, \, \tilde{y}, \, 0, \, 0 \right]^\intercal$
    \State $\state \gets \left[ \state, \, \bm{\xi}_i \right]$
    \EndFor
  \end{algorithmic}
  \caption{Adding new clauses $\clauses_\text{new}$ to a spatial model}
  \label{alg:add_new}
\end{algorithm}

A new spatial model is constructed when a set of new grammar clauses $\clauses_\text{new}$ is received by the system, with the previous spatial model used as the starting system state $\mathbf{x}_0$. All new point-masses are first given an initial state through the iterative initialisation procedure shown in Algorithm \ref{alg:add_new}. The procedure sorts all new point-masses in descending order ranked by the total weight of constraints on their position (i.e. sum of $K$ values for all springs attached to the point-mass), then iteratively places each point-mass at the position $(\tilde{x}, \tilde{y})$ that most satisfies the constraints imposed by the natural length of all attached springs. The ordering ensures point-masses whose positions are most heavily constrained by springs are placed first when those constraints are most likely to be satisfiable. Each point-mass is given zero starting velocity when placed.

After a spatial model is augmented with new clauses, a new spatial model is created using the updated system motion model. The function modelling the system motion, for a system with $m$ point-masses and $n$ springs, is defined as:
\begin{equation}
  \fnCon(t, \state) = \sum_{j=1}^{n} \fnSpr(\state,\springParam_j) + \sum_{i=1}^{m} \fnDam(\state,\pmParam_i) + \sum_{i=1}^{m} \fnExp(\state,\fnCOM(t),\pmParam_i)
  \label{eq:system_configuration}
\end{equation}
where $\fnSpr(\state,\springParam_j)$ is the force applied by spring $j$ with parameters $\springParam_j$, given system state $\state$. $\fnDam(\state,\pmParam_i)$ is the viscous friction force on point-mass $i$ with parameters $\pmParam_i$, given system state $\state$. The viscous friction ensures a solution is reached by introducing damping motion. $\fnExp(\state,\fnCOM(t),\pmParam_i)$ adds an expansion force pushing point-mass $i$ away from $\fnCOM(t)$, where $\fnCOM(t)$ is the centre of explored mass in the robot's underlying metric map (see Fig. \ref{fig:system_outline}). The expansion component encourages the spatial model to expand away from already explored spaces, particularly when a place's location estimate is underconstrained (e.g. when a place has a spring suggesting relative distance but no spring constraining direction).

The system dynamics are simulated using iterative Runge--Kutta integration of the ordinary differential equation $\fnCon(t,\state)$, until the settling criteria $\fnSet(\dstate)$ is met. Once the settling criteria is met, an imagined spatial model is returned as the positions of the system's point-masses. The settling criteria is defined as:
\begin{equation}
  \begin{split}
    \fnSet(\dstate) = \sqrt{\ddot{x}_i^2 + \ddot{y}_i^2} < \setLa \quad &\land \quad \sqrt{\dot{x}_i^2 + \dot{y}_i^2} < \setLv \\
    \forall \enspace i=1, &\dots, m
    \label{eq:settling_criteria}
  \end{split}
\end{equation}
where $\setLa$ is the acceleration threshold at which a point-mass is deemed to be settled, and $\setLv$ is the velocity threshold. The two conditions combine to continue simulating the system while any point-mass is in motion, or accelerating due to unbalanced forces. System dynamics cycle energy between spring tension and point-mass motion as they explore possible layouts for the imagined spatial model. Friction drives the system to a minima where motion ceases, denoting the most representative layout. The graph shown in the imagination phase of Fig. \ref{fig:example_process} depicts the total energy over system time. To highlight the subtle modelling differences between spatial layout and spatial hierarchy, the final imagined spatial models are shown split by relational clause type in the bottom right of the figure.

\subsection{Reconciling Imagination with Observation through Locational Clauses}

Sweeping assumptions of distances, scales, sizes, and directions are made in the spatial model to imagine spatial layouts solely from symbols, and these will likely conflict with the robot's observations of its environment. No single set of assumptions apply to every built environment; assumptions must be adapted for differences in scale, structure, and between indoor or outdoor environments. The link between symbols and the real world environment in locational clauses provides information that can help inform these assumptions. We use this information to align the imagined spatial models in the abstract map with reality, and refine assumptions through experience. This work only incorporates locational clauses for the robot's frame of reference (see \cite{schulz2015robot} for an approach that could be adapted to enhance the system described below).

When conflicts between imagination and reality occur, they are reconciled in the spatial model by trusting observations over what has been imagined solely from symbols. To exert authority in the imagined spatial model two tools are employed: fixing point-masses and changing spring stiffness. Upon observing a cue at $({}^\refframe{W}\posx,{}^\refframe{W}\posy)$ describing the relative distance and direction to $\toponym \in \tops$ as $\dist \in \{\varnothing,\real\}$ and $\dir \in \{\varnothing,\angles\}$ respectively, a fixed point-mass is added at $({}^\refframe{W}\posx,{}^\refframe{W}\posy)$ in the malleable spatial model. Here $\varnothing$ is used when a cue doesn't communicate distance or direction (for example a sign with only an arrow gives no $\dist$ value). Springs are added between point-mass $\pmParam_{\iota(\toponym)}$ and the fixed point-mass, with a high stiffness coefficient. The high stiffness means springs created from observations will override suggestions from springs created by loosely imagining spatial layout solely from symbols.

The robot's observations can also be used to update assumed scales in the abstract map, and exploit refined values for improved imagination. When the model was first imagined, the natural lengths of distance springs were set based on estimates of environment scale. With the benefit of real world observations, scaling factors can be manipulated to improve the abstract map's earlier estimates.

Scaling factors $\factor^a_b$ are employed for each unique unordered pair of levels $(a,b) \in (\posints)^2$ in the hierarchy graph. For instance, the example in Fig. \ref{fig:example_process} has three levels with level 1 corresponding to rooms, level 2 to buildings, and level 3 to university campuses. The hierarchy creates six scaling factors ($\factor^1_1$, $\factor^1_2$, $\factor^1_3$, $\factor^2_2$, $\factor^2_3$, and $\factor^3_3$), where $\factor^1_1$ corresponds to the average distance between adjacent rooms, $\factor^2_2$ the average distance between adjacent buildings, $\factor^1_2$ the average distance between any room within a building, etc. Scaling factors between two hierarchy levels are given a default value until a distance $r_o$ is observed between the levels. $r_o$ is obtained when labels for both endpoints of a distance spring have been observed. Once a distance has been observed, the spatial model uses a scaling factor instead of the default value.

Scaling factors are calculated by comparing the observed length $r_o$ of springs with their initial estimated natural length $\parXr$. A scaling factor for hierarchy levels $(a,b)$ is the weighted arithmetic mean of the observed scaling error ($r_o / \parXr$) for the $n$ distances observed between toponyms in $a$ and $b$:
\begin{equation}
  \factor^a_b = \frac{\sum\limits_{i = 1}^{n} K_i \frac{r_{o_i}}{r_{n_i}}}{\sum\limits_{i=1}^{n} K_i}
  \label{eq:scaling_factors}
\end{equation}
where the stiffness $K_i$ is used as the weight. The effect of the scaling factors can be seen in the hierarchical springs and spatial model created in Fig. \ref{fig:example_process}, and the results shown in Section \ref{subsec:results_robot}.

Lastly, an exploration scaling factor $\explorefactor$ is applied to the natural length of each distance spring when a goal is not found at its imagined location. This factor expands the scope of exploration in larger than expected sections of environments, like outdoor environments with less repetitive structure. $\explorefactor$ has an initial value of $1$ and is increased multiplicatively by an exploration step $\explorestep$ when a goal is not found where it is expected. Step increases are applied until the robot observes a new navigation cue. The increases in $\explorefactor$ expand the spatial model, encouraging the robot to search for the goal outside of already visited areas. Upon observing a new navigation cue, $\explorefactor$ is reset to 1 and the normal process is resumed.

\section{Experimental Process}
\label{sec:procedure}

There is relatively little prior work on symbolic navigation of unseen places and no relevant benchmarks for evaluating navigation performance. This section describes our approach to performance evaluation in real world built environments. Human participants, with the symbolic navigation abilities that motivated this research, were used as a performance baselines. The symbolic navigation task used for evaluation was one that is a common part of the human symbolic navigation experience: finding an animal at the zoo. The research was approved by the QUT Human Research Ethics Committee (approval number 1800000392).

\begin{figure}[p]
  \centering
  \input{./figs/experiment_maps.tex}
  \caption{Spatial and hierarchical maps of the fictional zoo used for both the human and robot studies.}
  \label{fig:experiment_maps}
\end{figure}

\begin{figure}[t]
  \centering
  \input{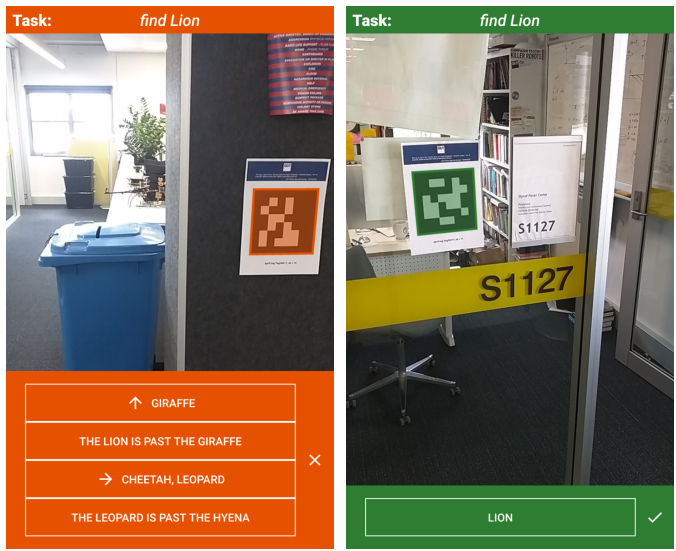}
  \caption{Screenshots from the mobile application human participants used to detect AprilTags. Each detection is highlighted, and the decoded symbolic spatial information displayed.}
  \vspace{\shrinkfactor}
  \label{fig:experiment_app}
\end{figure}

We used a fictional zoo environment for the experiments, with animal enclosures grouped into five themed areas branching off the ``Zoo Foyer'' as shown in Fig. \ref{subfig:experiment_map}. The zoo environment encompassed the entire floor of a university campus building. To level the playing field for humans and robots, all navigation cues (place names and direction boards) were encoded in AprilTags \cite{wang2016apriltag} which were physically placed in the environment. AprilTags employed a combination of text and arrows to emulate labels, natural language descriptions, directional signs, and signboards (examples can be seen in Fig. \ref{fig:experiment_app}). Symbolic spatial information was encoded in the AprilTags through a single static mapping for all trials. Places in the environment and navigation cues purposely had no visual resemblance of what they were representing (animals and AprilTags respectively). This removed insights like ``this looks like an aviary'', ``that looks like a Giraffe over in the far corner'', using contextual knowledge to ignore irrelevant environment text, and long-distance cue recognition---all of which are outside the scope of this research.

Each symbolic navigation task started outside the zoo, near the ``Exit'', and was deemed complete upon observation of the symbolic goal's label. The experiment consisted of 50 trials with 25 human participants completing a single navigation task each, and the robot completing 25 tasks starting each trial with no prior navigation knowledge. Human participants were aged between 18 and 59, with university education either completed or in progress, and had never previously visited the experiment environment. Trials were split into five unique navigation goals (``Lion'', ``Kingfisher'', ``Polar Bear'', ``Anaconda'', and ``Toilets'') with attempts from five human participants and five from the robot for each goal.

Participants were also provided with a graph of the zoo's spatial hierarchy (shown in Fig. \ref{subfig:experiment_graph}). For humans this was a printed sheet, whereas the robot used the graph to create relational clauses that were preloaded into the abstract map. The spatial hierarchy graph normalised contextual knowledge between participants, and removed discrepancies in contextual interpretations like whether the ``Cockatoo'' would be in the ``Bird Aviary'' or ``Outback Adventure''. Tools were given to both human and robot participants to read the information encoded in AprilTags. A purpose-built mobile phone application was provided to human participants as shown in Fig. \ref{fig:experiment_app}, and the robot employed a detector monitoring images from a panoramic camera.

Distance travelled was the performance measure used for both robot and human trials, with audio also recorded in human trials. In human trials, the path travelled was recorded manually on a map and directly from the robot's raw odometry data during robot trials. A fair basis for comparison was established by retracing the paths recorded for human trials with the same robot used in the robot trials. Audio was recorded during each human trial and in a brief post-interview where discussions were guided through three topics: describing navigation experiences, exploring what guided navigation (and if cues besides AprilTags played a role), and comparing AprilTag cues to the human navigation cues typically found in built environments.

We designed the experiment to maximise the validity of the comparison between robot and human performance. Additional experimental controls included ensuring the robot and human were given the same contextual knowledge via the graph in Fig. \ref{subfig:experiment_graph}, limiting the AprilTag detection range in the mobile phone application to match the robot's \SI{4}{\metre} detection range, and requiring human participants to have never previously visited the experimental environment.

\subsection{Parameterisation in the Abstract Map}

For the robot trials, a number of parameters controlled the imagination of spatial models and incorporation of symbol observations. The parameters are listed below, with notes about their selected values:
\begin{itemize}
  \item Each preposition $s$ is hand-mapped to a set of springs with minimal value tuning of parameters required (only four $\parXt \in \{\pm\pi,\pm\pi/2\}$ and two $\parXr \in \{1,0.5\}$ values were used across all preposition interpretations).
  \item The system used five different stiffness values, $K \in \{2.5,1,0.5,0.1,0.01\}$ ($2.5$ was reserved for observation springs attached to fixed point-masses, with the remaining values used in preposition to spring conversion).
  \item All point-masses had a mass of \SI{1}{\kilogram}.
  \item A viscous friction coefficient of $0.1$ was used, with higher values introducing unnecessary overshoot in spring motion and increasing settling time.
  \item $0.01$ was the expansion coefficient used to scale the proportional relationship between distance from centre of explored mass and force in $\fnExp(\state,\fnCOM(t),\pmParam)$ (larger values caused the spatial model to stretch).
  \item $\setLa$ and $\setLv$ were both set to $0.1$. Increasing the values caused the imagination phase to finish before point-masses have finished moving, whereas low values resulted in delayed detection of motion completion.
  \item The zoo hierarchy had three levels, with rooms, themed areas, and zoos corresponding to levels 1-3 respectively. Scaling factors were given the following default starting values: $\factor^1_1 = \SI{4}{\metre}$, $\factor^1_2 = \SI{5}{\metre}$, $\factor^1_3 = \SI{20}{\metre}$, $\factor^2_2 = \SI{15}{\metre}$, $\factor^2_3 = \SI{15}{\metre}$, and $\factor^3_3 = \SI{50}{\metre}$.
  \item A $25\%$ exploration step was used to rapidly expand imagined location estimates when goals were not found.
\end{itemize}

\subsection{Robot Configuration}

An Adept GuiaBot mobile base was used in the robot experiments, with panoramic images from a 360\textdegree{} Occam camera scanned for AprilTags. The standard ROS navigation stack provided the SLAM and spatial navigation components from Figure {\ref{fig:system_outline}}, with the robot controlled by pose goals produced from the abstract map.

\section{Results} \label{sec:results}

Symbolic navigation performance with the abstract map was evaluated against human participants, with a number of qualitative insights gained about the human symbolic navigation process. This section provides a quantitative comparison of symbolic navigation performance between a robot navigation system employing the abstract map and human participants, expanded details describing one robot and one human trial, as well as a qualitative summary of insights from the human participants. Numbers in the text such as \#12 refer to the tag numbers shown in Fig. \ref{subfig:experiment_map}.

\subsection{Robot Performance against a Human Benchmark}

\begin{table}[t]
  \centering
  \renewcommand{\arraystretch}{1.5}
  \begin{tabular}{@{}lccc@{}}
    \toprule
    Symbolic Goal & Human (\si{\metre}) & Robot (\si{\metre}) & Improvement (\%) \\ \midrule
    Kingfisher             & $34.4$             & $36.9$             & \textcolor{red!80!black}{$-7.3$}                      \\
    Toilets                & $52.1$             & $37.6$             & \textcolor{green!50!black}{$27.8$}                     \\
    Lion                   & $49.4$             & $45.6$             & \textcolor{green!50!black}{$7.6$}                      \\
    Polar bear             & $57.6$             & $40.5$             & \textcolor{green!50!black}{$29.8$}                     \\
    Anaconda               & $38.7$             & $38.9$             & \textcolor{red!80!black}{$-0.6$}                      \\
    \textbf{Overall}                & $\bm{46.4}$             & $\bm{39.9}$             & \textcolor{green!50!black}{$\bm{11.5}$}                     \\ \bottomrule
  \end{tabular}
  \caption{Average distance travelled for the human benchmark $\bar{x}_h$, and robot trials $\bar{x}_r$ (improvement is $1-\bar{x}_r/\bar{x}_h$)}
  \label{tab:results}
\end{table}

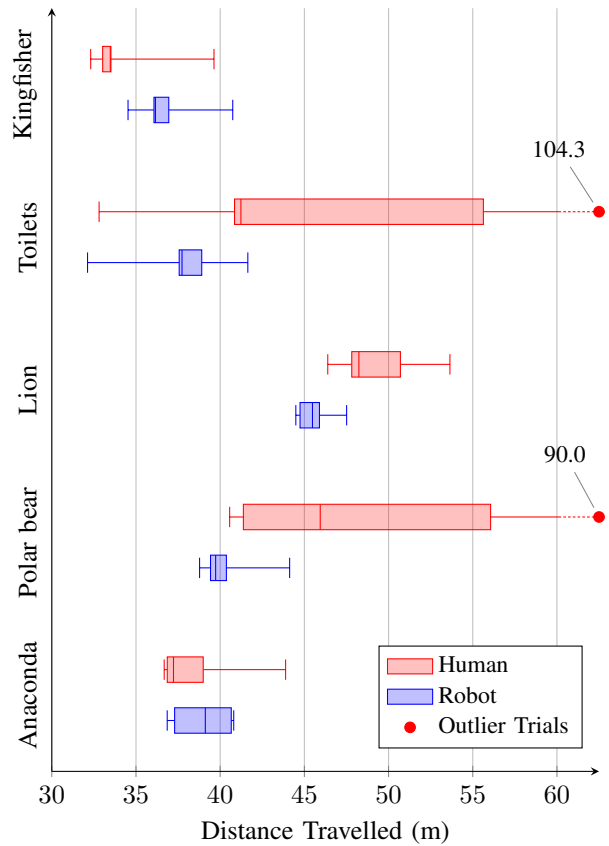
\begin{figure}[tp]
  \centering
  \begin{tikzpicture}
  \begin{axis}[
      boxplot/draw direction=x,
      boxplot/whisker range=1000,
      boxplot/box extend=0.5,
      xtick pos=left,
      ytick pos=left,
      axis y line=left,
      axis x line=bottom,
      xmajorgrids=true,
      width=\columnwidth,
      height=1.325\columnwidth,
      xlabel={Distance Travelled (m)},
      xmin=30,
      xmax=62.5,
      ymin=0,
      ymax=15,
      ytick style={draw=none},
      ytick={13.5,10.5,7.5,4.5,1.5},
      yticklabels={Kingfisher,Toilets,Lion,Polar bear,Anaconda},
      y tick label style={rotate=90},
      every axis plot/.append style={fill,fill opacity=0.25},
      legend pos=south east,
      legend cell align=left
    ]

    \addplot+ [boxplot={draw position=14}, area legend, red] table[row sep=newline, y index=0] {
      32.31002160295536
      33.511470964615
      33.02154990654171
      33.473877641130414
      39.63258819856321
    };
    \addplot+ [boxplot={draw position=13}, area legend, blue] table[row sep=newline, y index=0] {
      36.06795128583621
      36.94700814519428
      40.74612069270048
      34.527517081095
      36.15554173789272
    };
    \addplot+ [boxplot={draw position=11}, forget plot, red] table[row sep=newline, y index=0] {
      32.80904736033748
      89.95978921173769
      40.85287167086429
      55.63106007968242
      41.231169840037175
    };
    \addplot+ [boxplot={draw position=10}, forget plot, blue] table[row sep=newline, y index=0] {
      41.64234647246752
      32.123313079628375
      37.56488837511367
      38.90493291919638
      37.733763337986304
    };
    \addplot+ [boxplot={draw position=8}, forget plot, red] table[row sep=newline, y index=0] {
      53.644638189313596
      47.814369028078794
      48.23869591684716
      50.70377533981668
      46.389597105115115
    };
    \addplot+ [boxplot={draw position=7}, forget plot, blue] table[row sep=newline, y index=0] {
      45.895081471566996
      47.5104669851718
      44.7423996452446
      45.48094913218878
      44.492758092006895
    };
    \addplot+ [boxplot={draw position=5}, forget plot, red] table[row sep=newline, y index=0] {
      41.377022038565784
      40.56260600506652
      104.2930621312444
      45.94183023431989
      56.06052830938193
    };
    \addplot+ [boxplot={draw position=4}, forget plot, blue] table[row sep=newline, y index=0] {
      40.36846307885318
      38.77675330202988
      39.42682275063831
      44.12503906821261
      39.73139296867467
    };
    \addplot+ [boxplot={draw position=2}, forget plot, red] table[row sep=newline, y index=0] {
      37.22718914324148
      36.67537433084827
      43.88541779585317
      38.99250058688703
      36.857056169764554
    };
    \addplot+ [boxplot={draw position=1}, forget plot, blue] table[row sep=newline, y index=0] {
      40.80502767743813
      40.658245100117426
      36.85344711943129
      37.2946943630243
      39.117223879125206
    };
    \addplot [densely dotted, white, thick, forget plot] coordinates {
      (60.25, 5)
      (62.25, 5)
    };
    \addplot [densely dotted, white, thick, forget plot] coordinates {
      (60.25, 11)
      (62.25, 11)
    };
    \addplot [only marks, mark=*, red, fill opacity=100] coordinates {
      (62.5, 5)
      (62.5, 11)
    } node[pos=0.0, pin={[black]95:{\small 90.0}}]{} node[pos=1.0, pin={[black]95:{\small 104.3}}]{};
    \legend{\small Human, \small Robot, \small Outlier Trials}
  \end{axis}
\end{tikzpicture}
  \caption{Human and robot performance in the experimental trials, measured in distance travelled. Single outlier results occurred in the human ``Toilets'' and ``Polar bear'' trials, as noted in the graph.}
  \vspace{\shrinkfactor}
  \label{fig:experiment_results}
\end{figure}

Table \ref{tab:results} compares the mean distances travelled human and robot participants for each of the five symbolic navigation tasks. Fig. \ref{fig:experiment_results} summarises the distances travelled each of the 50 trials. Human participants travelled an average distance of \SI{46.4}{\metre} whereas the robot travelled \SI{39.9}{\metre} on average, with the abstract map guiding the robot to more efficient task completion in three of the five navigation tasks. Overall, the abstract map guided the robot to complete tasks \SI{11.5}{\percent} more efficiently than human participants, and \SI{5.3}{\percent} more efficiently with the two outlier human results removed.

\subsection{Expanded Robot Result: Finding the Lion}
\label{subsec:results_robot}

\begin{figure}[!hbp]
  \centering
  \includegraphics[width=\columnwidth]{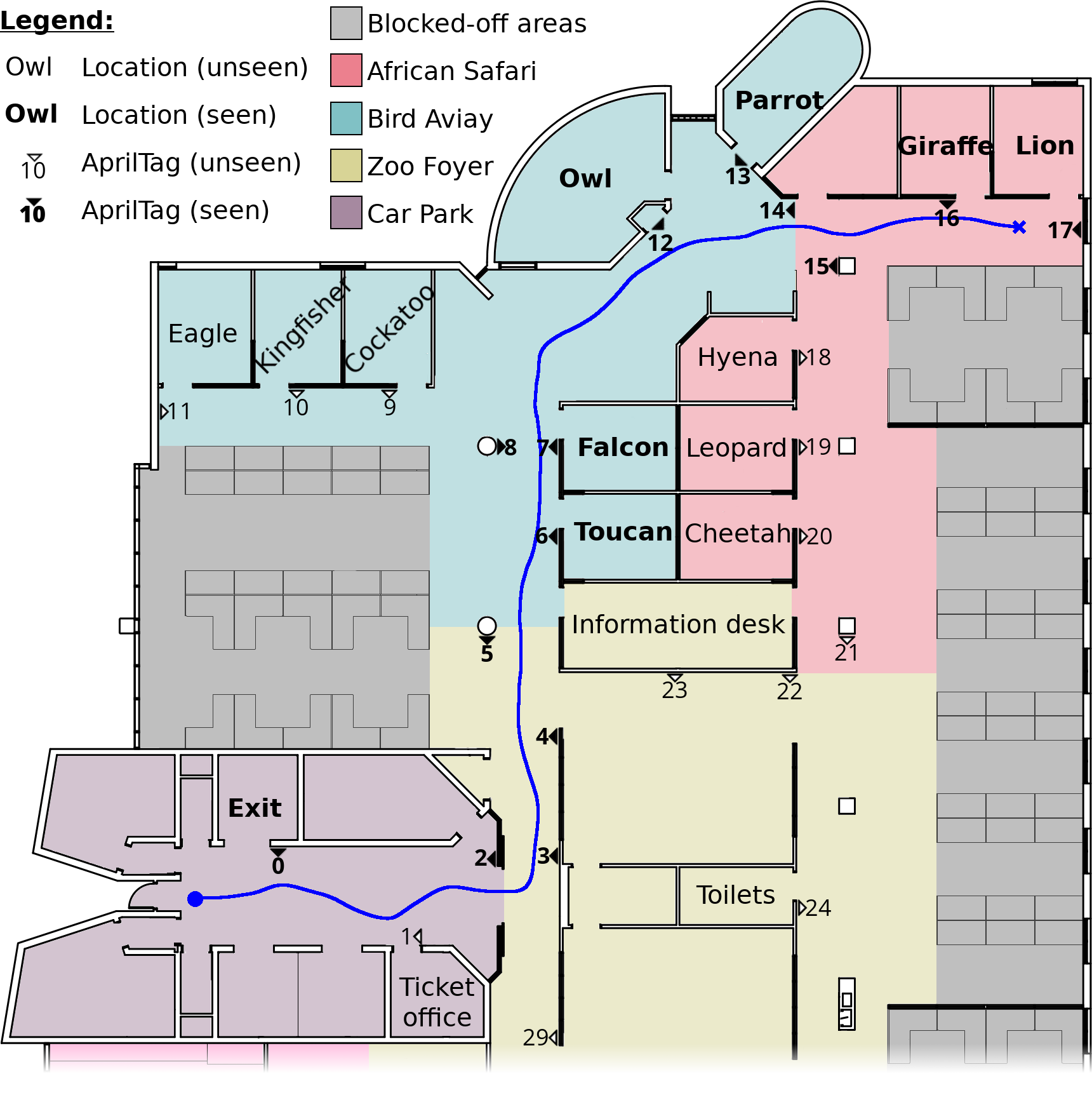}
  \caption{``Find the lion'', robot trial number 5. The robot started at the circle, and found the ``Lion'' at the cross. Tags and locations observed by the robot are shown in bold.}
  \vspace{\shrinkfactor}
  \label{fig:lion_robot}
\end{figure}

Fig. \ref{fig:lion_robot} shows the full path taken in the shortest robot trial (\SI{44.5}{\metre}) for the symbolic navigation task to ``find the lion''. The robot near the ``Exit'', with no existing map of the world and no prior information describing the zoo. It built a metric map as it travelled through the environment using a SLAM system, using straight line navigation plans when planning in unseen spaces. Examples of the underlying SLAM system are shown in Figure \ref{fig:experiment_abstract_map}. An initial spatial model for the zoo was imagined using relational clauses extracted from the zoo hierarchy graph shown in Fig. \ref{subfig:experiment_graph}. With no symbolic spatial information describing the zoo layout, the system began moving to its imagined location for the ``Lion'' as shown in Fig. \ref{subfig:experiment_abstract_map_a}.

\begin{figure*}[p]
  \centering
  \input{./figs/experiment_abstract_map.tex}
  \caption{Process undertaken by a robot navigation system using the abstract map to successfully navigate the robot to the ``Lion'' (see \url{https://btalb.github.io/abstract_map/} for videos of the process).}
  \label{fig:experiment_abstract_map}
\end{figure*}

While avoiding obstacles and following the path planned by the underlying navigation system, the robot continued moving through free space towards the abstract map's current imagined location for the goal (near the ``Information Desk'' in reality). The robot proceeded into the ``Zoo Foyer'' upon seeing the label, and observed the signboard in tag \#3 (see Fig. \ref{subfig:experiment_signs} for signboard contents). Using the wealth of directional information in the signboard, the abstract map was heavily refined as shown in Fig. \ref{subfig:experiment_abstract_map_b} and guided the robot left in search of the ``Lion''.

Next, the robot passed tag \#4 which communicated that the ``African Safari is past the Information Desk'' and the ``Information Desk'' was to the right. The abstract map was updated with the information, but suggested a location for the ``Lion'' in between going right and straight ahead at the fork due to also having information from tag \#3 suggesting the ``African Safari'' was straight ahead. With the underlying path planning navigation system choosing to go straight as shown in Fig. \ref{subfig:experiment_abstract_map_c} (right was also chosen in other trials), the robot proceeded through the ``Bird Aviary'' and past the labels for the ``Toucan'' and ``Falcon''. Tag \#8 communicated a number of directional messages regarding the remaining birds in the ``Bird Aviary'' at the end of the junction. Importantly, it communicated that the ``Owl'' was to the right and the ``African Safari is past the Owl'', causing the abstract map's estimate to guide the robot right at the junction in search of the ``African Safari'' as shown in Fig. \ref{subfig:experiment_abstract_map_d}.

After passing labels for the ``Owl'' and ``Parrot'', the robot found a label for the ``African Safari'' at tag \#14 and signboard describing the safari at tag \#15. Amongst other information, the signboard communicated that the ``Giraffe'' was directly ahead and the ``Lion was past the Giraffe'', which the was used in the abstract map to guide the robot straight ahead as seen in Fig. \ref{subfig:experiment_abstract_map_e}. A final update of the abstract map's spatial model was performed upon observing the ``Giraffe'' label, before the robot completed the symbolic navigation task by finding the label for the ``Lion'' (as seen in Fig. \ref{subfig:experiment_abstract_map_f}).

\subsection{Expanded Human Result: Finding the Lion}

\begin{figure}[t]
  \centering
  \includegraphics[width=\columnwidth]{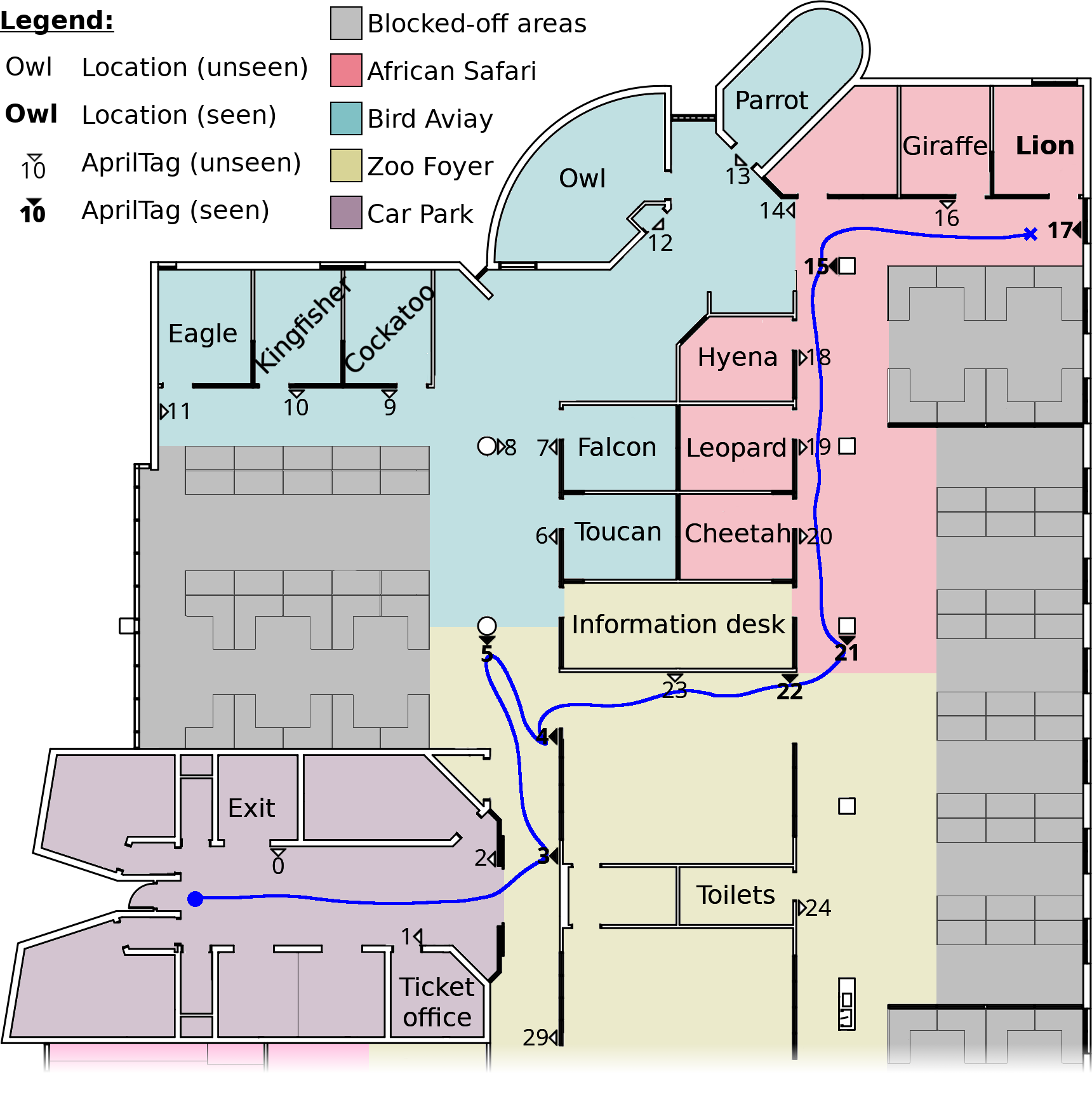}
  \caption{``Find the lion'', human trial number 4. The human started at the circle, and found the ``Lion'' at the cross. Tags and locations observed by the participant are shown in bold.}
  \vspace{\shrinkfactor}
  \label{fig:lion_human}
\end{figure}

The fourth best human participant (\SI{50.7}{\metre}) began at the same location as the robot, as shown in Fig. \ref{fig:lion_human}, and was instructed to ``find the lion''. The participant, who had never been to the environment before, was given a mobile phone with the application shown in Fig. \ref{fig:experiment_app}, and the graph of the zoo hierarchy shown in Fig. \ref{subfig:experiment_graph}. Labels for the ``Exit'', ``Ticket Office'', and ``Zoo Foyer'' were missed by the participant as they walked directly to the main signboard at tag \#3, which had directional labels for nine different locations (including most of the themed animal areas). The participant spent time processing and double checking the information before proceeding left.

Next, the participant walked directly past the signboard in tag \#4 to observe the label for the ``Bird Aviary'' at tag \#5. The participant then backtracked to find the signboard in tag \#4, later commenting that seeing the ``Bird Aviary'' label was ``negative information'' that made them question their current approach. In looking for the ``African Safari'', the participant proceeded directly past the label for the ``Information Desk'' at tag \#23. A directional signboard and label for the ``African Safari'' was observed at tags \#21 and \#22 respectively, with the participant proceeding down the hallway of the ``African Safari'' while deliberately not scanning tags \#18--\#20 on the left wall. At the end of the walkway they observed the signboard from tag \#15, and then observed the ``Lion'' label after purposely walking past the label for ``Giraffe''.

In the post-interview, the participant shared their navigation process. When queried about the skipped AprilTags, the participant described not noticing the tags in the ``Carpark'' and initially at tag \#4 as well as ``guessing'' based on the structure of the environment. Time spent at the main signboard in tag \#3 was described by the participant as ``reading it twice to try and remember it'' before describing trying ``to picture'' what the sign was communicating. To picture what the sign was communicating, the participant described ``trying to picture it as a route, or left and right [forks with a focus on] which directions to go'' while using the environment to eliminate unrealistic routes like ``requiring you to go through chairs or where people are working''. The comment suggested a symbolic navigation process involving an axial structure like a simple topological graph, but further questions revealed no more details. The trial concluded with AprilTags described as ``pretty similar'' to typical navigation cues, aside from the minimal potential for confusion in interpreting arrows relative to the phone screen rather than AprilTag placement in the real world.

\subsection{Insights from Human Participants}

Human participants offered a number of insights through verbalisation while completing the navigation tasks, and question responses in the post-interview. Listed below are the insights that were deemed relevant to evaluating study outcomes and understanding the human navigation process in unseen built environments:

\begin{itemize}
  \item No participants believed using tags instead of normal navigation cues affected their navigation process (\SI{88}{\percent} said it was comparable, with the remaining participants offering no direct answer).
  \item The majority (\SI{76}{\percent}) of participants commented on the ``added hassle'' in reading a tag with the phone rather than simply reading a normal cue, affirming the study limited the efficacy of human navigation cue perception to a level comparable with the robot.
  \item Participants often went out of their way to find the next AprilTag cue (as stated by \SI{60}{\percent}), generally commenting that they ``relied on the cues or tags more than their best guess'' of where the goal could reside. The abstract map has the opposite approach to the problem: it primarily follows its imagined goal location, and updates the location with any cues it observes along the way (an approach only mentioned by \SI{12}{\percent} of participants).
  \item \SI{32}{\percent} of participants walked directly past cues even though they were placed in conspicuous places at eye level. Most claimed to not notice the AprilTag, suggesting that visual attention may play a part in human navigation performance.
  \item Participants identified a wide range of cues used for navigation besides the AprilTags, grouped under three categories: visual, environmental context, and deeper cognition (present in \SI{56}{\percent}, \SI{32}{\percent}, and \SI{16}{\percent} of responses respectively). Visual cues from the environment included walkways, physical spatial structure, and lack of typical zoo features (e.g. thematic elements in the ``Artic Frontier''). Environmental context cues included knowing which areas were out of bounds, labels being more likely to be on offices, signboards more likely at choice points, and likely segmentations of space for the zoo areas. Lastly, deeper cognitive cues were employed like matching the spaces to where it looked like there was enough room for all of the animals in the zoo hierarchy graph, and guessing the experiment designer's thought process.
  \item The navigation strategies employed by participants displayed a lot of variety. Strategies included trusting tags over instincts (\SI{56}{\percent}), using tags heavily at navigational choice points (\SI{40}{\percent}), deliberately walking past cues to trust instincts or guessing what was likely being communicated (\SI{24}{\percent}), exploratively wandering until feeling lost then looking for cues (\SI{12}{\percent}), using negative information in cues to rule out possible options (\SI{32}{\percent}), and applying the context of typical zoo layouts from past experiences (\SI{16}{\percent}).
  \item \SI{20}{\percent} of participants took paths that appeared to have no explanation, with \SI{8}{\percent} taking significant detours (the outlier results). Comments by participants suggested this was due to misunderstanding cues, failing to see locations within the physical environment, and employing intuition without any other guidance.
\end{itemize}

\section{Conclusions \& Future Work}
\label{sec:conclusion}

The study results demonstrated that the dynamics-based abstract map empowers traditional robot navigation systems with symbolic navigation performance comparable to humans in unseen built environments. This level of performance was achieved by first creating a speculative spatial model imagined from symbolic descriptions of places, then iteratively refining and improving that model with symbolic observations gained by the robot as it traverses the real world. A robotics-oriented grammar, specifically designed for the task of representing the symbolic spatial information embedded in navigation cues, allowed the abstract map to seamlessly incorporate the symbols implicit in navigation cues in built environments. Consequently, a viable approach to grounding human spatial symbols was demonstrated with the abstract map by using navigation cues in a robot navigation process.

From the results of the study---particularly the insights provided by the human participants---the following avenues for enhancing the abstract map are proposed:

\begin{itemize}
  \item Gaining a deeper understanding of the differences between abstract map and human performance by testing in larger-scale environments like buildings or campuses, with varying levels of symbolic spatial information;
  \item Enabling the abstract map to employ negative information (the sensation of ``this doesn't seem right'' felt when cues in the environment aren't describing what is expected); and
  \item Using long range cue detection to guide the underlying robot navigation process as done by human participants, rather than blindly moving to the goal and accepting cue observations along the way.
\end{itemize}

In this paper we have presented the abstract map, a system that allows a robot to enhance its navigation process with the rich information provided by symbols. By employing a spatial model based on multi-body dynamics, and utilising the symbolic spatial information embedded in an environment's purposefully placed navigation cues, this paper has demonstrated that a robot using the abstract map is capable of performing symbolic navigation at a level comparable to human performance. The method presented allows robots to move out of seen spaces described by limited subsets of human symbols and into real world human environments like schools, hospitals, offices, and zoos; an imperative transition in realising ambitions for robots to becoming ubiquitous co-inhabitants of built environments.

\section{Acknowledgements}

This research was supported under the Australian Research Council's Discovery Projects funding scheme (project number DP140103216).

We would also like to acknowledge the contributions of Ben Upcroft and Ruth Schulz to the early stages of this research.

\bibliographystyle{IEEEtran}
\bibliography{bib/google,bib/custom}

\end{document}